%% file: fracbits.tex
\def\year{2021}\relax
\documentclass[letterpaper]{article} 
\usepackage{aaai21}  
\usepackage{times}  
\usepackage{helvet} 
\usepackage{courier}  
\usepackage[hyphens]{url}  
\usepackage{graphicx} 
\def\UrlFont{\rm}  
\usepackage{natbib}  
\usepackage{caption} 
\usepackage{amsmath,amssymb}
\usepackage{subcaption}
\usepackage{hhline}
\usepackage{makecell}
\usepackage{multirow}
\usepackage{pifont}
\newcommand{\cmark}{\ding{51}}%
\newcommand{\xmark}{\ding{55}}%

\title{FracBits: Mixed Precision Quantization via Fractional Bit-Widths}

\input{tex_files/authors}

\begin{document}

\maketitle

\input{tex_files/abstract}
\input{tex_files/introduction}
\input{tex_files/related_work}
\input{tex_files/algorithm}

\input{tex_files/experiments}

\input{tex_files/conclusion}

\bibliography{bibfile}

\end{document}


\maketitle

\input{tex_files/supp}

\bibliography{bibfile}

%% file: tex_files/authors.tex
\author{

    Linjie Yang,\textsuperscript{\rm 1}\thanks{Equal Contribution}Qing Jin\textsuperscript{\rm 2}\protect\footnotemark[1]
}
\affiliations{

    \textsuperscript{\rm 1}US AI Lab, ByteDance Inc.\\

    linjie.yang@bytedance.com

    \textsuperscript{\rm 2}Northeastern University\\
    jinqingking@gmail.com
}

%% file: tex_files/abstract.tex
\begin{abstract}
Model quantization helps to reduce model size and latency of deep neural networks. Mixed precision quantization is favorable with customized hardwares supporting arithmetic operations at multiple bit-widths to achieve maximum efficiency. We propose a novel learning-based algorithm to derive mixed precision models end-to-end under target computation constraints and model sizes. During the optimization, the bit-width of each layer / kernel in the model is at a fractional status of two consecutive bit-widths which can be adjusted gradually. With a differentiable regularization term, the resource constraints can be met during the quantization-aware training which results in an optimized mixed precision model. Our final models achieve comparable or better performance than previous quantization methods with mixed precision on MobilenetV1/V2, ResNet18 under different resource constraints on ImageNet dataset.
\end{abstract}

%% file: tex_files/introduction.tex
\section{Introduction}
Neural network quantization~\cite{pact,releq,sat,autoq,xnor,dq,haq,dnas,balancedq,dorefa} has attracted large amount of attention due to the resource and latency constraints in real applications.
Recent progress on neural network quantization has shown that the performance of quantized models can be as good as full precision models under moderate target bit-width such as 4 bits~\cite{sat}. Besides, customized hardwares can be configured to support multiple bit-widths for neural networks~\cite{adabits}. In order to fully exploit the power of model quantization, mixed precision quantization strategies are proposed to strike a better balance between computation cost and model accuracy. With more flexibility to distribute the computation budgets across layers~\cite{releq,sat,dnas}, or even weight kernels~\cite{autoq}, the quantized models with mixed precision usually achieve favorable performance than the ones with uniform precision.

Current approaches for mixed precision quantization usually borrow ideas from neural architecture search (NAS) literature. Suppose we have a neural network with each convolution layer consisting of $N$ branches where each branch is the quantized convolution with different bit-width. Finding the best configuration for a mixed precision model can be achieved by preserving a single branch for each convolution layer and pruning all other branches, which is conceptually equivalent to some recent NAS algorithms that aim at searching sub-networks from a supergraph~\cite{proxylessnas,enas,fbnet,snas}. ENAS~\cite{enas} and SNAS~\cite{snas} employ reinforcement learning (RL) to learn a policy to sample network blocks from a supergraph. ReLeQ~\cite{releq} and HAQ~\cite{haq} follow this footprint and employ reinforcement learning to choose layer-wise bit-width configurations for a neural network. AutoQ~\cite{autoq} further optimizes bit-width of each convolution kernel using a hierarchical RL strategy. ProxylessNAS~\cite{proxylessnas} and FBNet~\cite{fbnet} adopt a path sampling method to jointly learn model weights and importance scores of each operation in the supergraph. DNAS~\cite{dnas} directly reuses this path sampling methods and adds a regularization term proportional to the computation cost or model size, in order to discover mixed precision models with a good trade-off between computational resources and accuracy. Uniform Sampling (US)~\cite{uniform_sampling} is a similar method which uses uniform sampling to sample subnetworks from the supergraph in training and then searches for pruned or quantized models using evolutionary algorithm.
\input{tabs/tab1}

However, previous approaches on mixed precision quantization mostly directly adopts NAS algorithms and do not leverage specific properties of quantized models. Different from NAS and model pruning, the quantitative difference of weights and activations with similar bits is small. For example, choosing 4 or 5 bits for one weight matrix only generates around $7.4\%$ difference in value, assuming weights are uniformly distributed on $[0,1]$ with linear quantization scheme. Thus the transition from one bit to its neighboring bits can be considered as a differentiable operation with appropriate parameterization. Recently, DQ~\cite{dq} utilizes the Straight-Through Estimation~\cite{ste} to facilitate differentiable bit-switching by treating bit-width of each layer as continuous parameters. Here, we propose a new approach to treat the bit-widths as continuous values by interpolating quantized weights or activation values of its two nerighboring bit-widths. Such an approach facilitates an efficient one-shot differentiable optimization procedure of mixed precision quantization. By allocating differentiable bit-widths to layers or kernels, it can enable both layer-wise and kernel-wise quantization. A high-level comparison of our methods and previous mixed precision methods is shown in Table~\ref{table:compare}.


In summary, our contribution of this work is threefold.
\begin{itemize}

\item We propose a fractional bit-widths formulation which creates a smooth transition between neighboring quantized bits of network weights and activations, facilitating differentiable search in the layer-wise or kernel-wise precision dimension.
\item Our mixed precision quantization algorithm only needs one-shot training of the network, greatly reduces exploration cost for resource restrained tasks.
\item Our simple and straight-forward formulation is ready to be used for different quantization schemes. We showed superior performance than uniform precision approaches and previous mixed precision approaches on a wide range of model variants and with different quantization schemes.
\end{itemize}





%% file: tabs/tab1.tex
\begin{table}[tb]
\centering
\setlength{\tabcolsep}{3pt}
\footnotesize
\caption{A comparison of our approach and previous mixed quantization algorithms. Our method FracBits achieves one-shot differentiable search and supports both channel-wise and kernel-wise quatization.}
\begin{tabular}{lccccccc}

\hline
Methods                 &HAQ                   &ReLeQ                 &AutoQ                 &DNAS                  &US                    & DQ                    & Ours         \\ \hline
differentiable              & \xmark & \xmark & \xmark & \cmark & \xmark & \cmark & \cmark \\ \hline
one-shot                                        & \xmark & \xmark & \xmark & \xmark & \xmark & \cmark & \cmark \\ \hline
channel-wise & \cmark & \cmark & \cmark & \cmark & \cmark & \cmark & \cmark \\ \hline
kernel-wise & \xmark & \xmark & \cmark & \xmark & \xmark & \xmark & \cmark \\ \hline
\end{tabular}
\label{table:compare}
\end{table}

%% file: tex_files/related_work.tex
\section{Related Work}
{\noindent\bf Quantized Neural Networks} Previous quantization techniques can be categorized into two types. The first type named post-training quantization directly quantizes weights and activations of a pretrained full-precision model into lower bit~\cite{krishnamoorthi2018quantizing,nagel2019data}. This type of methods typically suffer from significant performance degeneration, as the training progress is ignorant of the quantization procedure. Another type of techniques named quantization-aware training is proposed to incorporate quantization into training stage. Early studies in this direction employ a single precision for the whole neural network. For example, DoReFa~\cite{dorefa} proposes to transform the unbounded weights into a finite interval to reduce undesired quantization error introduced by infrequent large outliers. PACT~\cite{pact} investigates the effect of clipping activations from different layers, finding the layer-dependence of the optimal clipping-levels. SAT~\cite{sat} investigates the gradient scales in training with quantized weights, and further improves model performance by adjusting weight scales.
As another direction, some work assigns different bit-widths to different layers or kernels, enabling more flexible computation budget allocation. The first attempts employ reinforcement learning technique with rewards from estimated memory and computational cost by formulas~\cite{releq} or simulators~\cite{haq}. AutoQ~\cite{autoq} modifies the training procedure into a hierarchical strategy, resulting in fine-grained kernel-wise quantization. However, these RL strategies needs to sample and train a large number of model variants which is very resource-demanding. DNAS~\cite{dnas} resorts to a differentiable strategy by constructing a supernet with each layer comprised by a linear combination of outputs from different bit-widths. However, due to the discrepancy between the search process and final configuration, it still needs to retrain the discovered model candidates. To further improve the searching efficiency, we propose a one-shot differentiable search method with fractional bit-widths. Due to the smooth transition between fractional bit-width and final integer bit-width, our method embeds the bit-width searching and model finetuning stages in a single pass of model training. Meanwhile, our technique is also orthogonal to Uniform Sample (US)~\cite{uniform_sampling} which trains a supernet by uniform sampling and searches good sub-architectures with evolutionary algorithm.

{\noindent\bf Network Pruning} Network pruning is an orthogonal approach to speed up inference of neural networks to quantization. Early work~\cite{han2015deep} compresses bulky models by learning connection together with weights, which produces unstructured connection in the final network. Later, structured compression by kernel-wise~\cite{luo2017thinet} or channel-wise~\cite{gordon2018morphnet,he2017channel,slimming,ye2018rethinking} pruning is proposed, where the learned architecture is more friendly with acceleration on modern hardware. As an example,~\cite{slimming} identifies and prunes insignificant channels in each layer by penalizing on the scaling factor of the batch normalization layer. More recently, NAS algorithms are leveraged to guide network pruning. \cite{autoslim} presents a one-shot searching algorithm by greedily slimming a pretrained slimmable neural network~\cite{slimmable}. \cite{atomnas} proposes a one-shot resource-aware searching algorithm using FLOPs as an L1 regularization term on the scaling factor of the batch normalization layer. We adopt a similar strategy to use BitOPs and model sizes as L1 regularization which are computed based on the trainable fractional bit-widths in our framework.

{\noindent\bf }

%% file: tex_files/algorithm.tex
\section{Mixed Precision Quantization}

\input{figs/fig1}

In this section, we will introduce our proposed method for mixed precision quantization. Our one-shot training pipeline involves two steps: bit-width searching and finetuning. We first introduce the implementation of fractional bit-width, and integration of the resource constraint in the searching process. After that, we introduce implementation of kernel-wise mixed precision quantization.

\subsection{Searching with fractional bit-widths}
In order to learn bit-widths dynamically in one-shot training, it is necessary to make them differentiable and define their derivative accordingly. To this end, we first examine a generic operation $f_k(x)$ that quantizes a value $x$ to $k$-bit. Typically, $f_k(x)$ is well-defined only for positive integer values of $k$. To generalize bit-width to an arbitrary positive real number $\lambda$, we apply first-order expansion around one of its nearby integer, and approximate the derivative at this integer by the slope of the segment joining the two adjacent grid points neighboring $\lambda$. Such a linear interpolation reads

\begin{equation}
\label{eq:fracbit}
    f_\lambda(x)\approx f_{\lfloor\lambda\rfloor}(x)+(\lambda-\lfloor\lambda\rfloor)(f_{\lceil\lambda\rceil}(x)-f_{\lfloor\lambda\rfloor}(x))
\end{equation}
where $\lfloor\cdot\rfloor$ and $\lceil\cdot\rceil$ denote the floor and ceiling function, respectively. In other words, we can approximate an operation with a fractional bit-width by a linear combination of two operations with integer bit-widths, thus naturally achieving differentiability on it and making it learnable through typical gradient-based optimization, such as SGD. Note that the approximation in~\eqref{eq:fracbit} turns into a strict equality if the original operation $f_k(x)$ is linear in $k$ or if $\lambda$ takes an integer value. The basic idea is illustrated in Fig.~\ref{fig:fracbit}. In~\eqref{eq:fracbit}, the two rounding functions floor and ceiling on bit-width has vanishing gradient with respect to the argument, and thus the partial derivative of~\eqref{eq:fracbit} with respect to $\lambda$ is given by

\begin{equation}
    \frac{\partial}{\partial\lambda}f_\lambda(x)=f_{\lceil\lambda\rceil}(x)-f_{\lfloor\lambda\rfloor}(x)
\end{equation}
The difference of such an linear interpolation scheme compared to the widely-adopted straight through estimation (STE)~\cite{ste} is that it uses soft bit-widths in both forward and backward propagation, rather than hard bit-widths in forward and soft bit-widths in back-propagation, as adopted by~\cite{dq}. In this way, the computed gradient reflects the true direction that the network parameters need to evolve along which results in better convergence.



Throughout we will adopt the DoReFa scheme for weight quantization, and the PACT scheme for activation quantization. The quantization function for both is the same, defined as
\begin{equation}
\label{eq:quant}
    q_k(x)=\frac{1}{a}\Big\lfloor ax\Big\rceil
\end{equation}
where $x\in[0,1]$, $\lfloor\cdot\rceil$ indicates rounding to the nearest integer, and $a$ equals $2^k-1$ where $k$ is the quantization bit-width. Thus, for both quantization, we have $f_k(x) = q_k(x)$ for integer bit-widths, and quantization with fractional bit-widths is implemented with Eq.~(\ref{eq:fracbit}). The weight quantization is given by $Q_W=2q_{\lambda_w}(\widetilde{W})-1$, where $\widetilde{W}$ is the transformed weight clamped to the interval $[0, 1]$; activation quantization is given by $Q_X=\alpha q_{\lambda_a}(\frac{X}{\alpha})$, where $\alpha$ is a learnable parameter and $X$ is the original activation clipped at $\alpha$. $\lambda_w$ and $\lambda_a$ are the learnable fractional bit-widths for weight and activation, respectively. Also, it is possible to privatize bit-width to each kernel, enabling kernel-wise mixed precision quantization, as discussed later.

During the earlier searching stage, the precision assigned to each layer or each kernel is still undetermined, and we want to find the optimal bit-width structure through training. By initializing each bit-width with some arbitrary value, we can use~\eqref{eq:fracbit} to quantize weights and activations in the model to fractional bit-widths. Meanwhile, this allows us to assign different bit-widths to different layers or even kernels, as well as to furnish separate precision for weight and activation quantization. During the training process, the model gradually converges to an optimal bit-width for both weight and activation corresponding to each unit, enabling quantization with mixed precision.

\subsection{Resource constraint as penalty loss}
Restricting storage or computation cost is essential for model quantization, as the original purpose of quantization is to save resource consumption when deploying bulky model on portable devices or embedded systems. To this end, previous work resort to constraining on different metrics during the optimization procedure, including memory footprints~\cite{dq}, model size~\cite{dq,haq}, BitOPs~\cite{uniform_sampling,dnas} and even estimated latency or energy~\cite{autoq,haq}. Here, we focus on model size in bits (Bytes) for weight-only quantization, and the number of BitOPs for quantization on both weight and activation, as they can be directly calculated from assigned bit-widths. Note latency and energy consumption~\cite{autoq,haq} may seem to be more practical measures for real applications. However, we argue that BitOPs can also be a good metric since it is solely determined by the model itself rather than different configurations of hardwares, simulators and compilers, which guarantees fair comparison between different approaches and advocates reproducible research.



Weight-only quantization targets at shrinking the model size, while floating point operation is still needed during inference. Model size are usually expressed in terms of the required number of bits to store weights (and bias) in the model. For a weight $w$ of $k_w$-bit, the size is simply $k_w$. The generalized model size for a fractional bit-width $\lambda_w$ is thus $\lambda_w$.
The size of the whole model can be obtained by summing over all weights in the model. Note that the bit-width can be shared among all weights in the whole layer or along each kernel (as discussed later), 
corresponding to layer-wise or kernel-wise quantization, respectively. For example, for a typical 2D convolution layer (without grouping) sharing the same fractional bit-width $\lambda_w$ among all weights, the size is given by $\lambda _wc_{in}c_{out}k_xk_y$, where $c_{in}$ is the number of input channels, $c_{out}$ is the number of output channels, and $k_x$ and $k_y$ represent the horizontal and vertical kernel sizes, respectively.

Quantization on both weights and activations can effectively decrease computation cost for real application, which can be measured with number of BitOPs involved in multiplications. Suppose a weight value $w$ and an activation value $a$ involved in multiplication are quantized to $k_w$-bit and $k_a$-bit, respectively. The number of BitOPs for such a multiplication is
\begin{equation}
    \mathrm{comp}_{wa}=k_wk_a
\end{equation}
This expression is bi-linear in $k_w$ and $k_a$, which means that for fractional bit-widths $\lambda_w$ and $\lambda_a$,~\eqref{eq:fracbit} leads to
\begin{equation}
    \mathrm{comp}_{wa}=\lambda_w\lambda_a
\end{equation}
The total computation cost of the model is the sum over all weights and activations. As for the example of 2D convolution layer, if all weights share the same fractional bit-width $\lambda_w$ and all input activations share the same fractional bit-width $\lambda_a$, the number of BitOPs is given by $\lambda_w\lambda_ac_{in}c_{out}k_xk_yo_xo_y$, where $o_x$ and $o_y$ represents the horizontal and vertical sizes of the output features, respectively.

{\noindent\bf Targeting prescribed objective} With constraints defined properly, we are able to penalize on them to enable constraint-aware optimization. Here, we directly define the penalty term as the L1 difference from some target constraint value by

\begin{subequations}
\begin{align}
    \mathcal{L}^\mathrm{size}&=\Big\lvert\sum_w\mathrm{size}_w-\mathrm{size}_t\Big\rvert\label{eq:loss_size}\\
    \mathcal{L}^\mathrm{comp}&=\Big\lvert\sum_{wa}\mathrm{comp}_{wa}-\mathrm{comp}_t\Big\rvert\label{eq:loss_comp}
\end{align}
\end{subequations}
where $\mathrm{size}_t$ and $\mathrm{comp}_t$ denote target constraints for model size and computation cost, respectively. The sum is taken over all weights in the model for model size constrained optimization, and is taken over all weights and all activations for computation cost constrained case. Following the convention adopted in most literature, for both constraints we only take into account those contributed by convolution and fully-connected layers.

Adding the penalty term to the original loss (such as cross entropy for classification task) with a coefficient $\kappa$, we arrive at the total loss for optimization
\begin{subequations}
\begin{align}
    \mathcal{L}^\mathrm{total}&=\mathcal{L}^\mathrm{cls}+\kappa\cdot\mathcal{L}^\mathrm{size}\\
    \mathcal{L}^\mathrm{total}&=\mathcal{L}^\mathrm{cls}+\kappa\cdot\mathcal{L}^\mathrm{comp}
\end{align}
\end{subequations}
It should be noted that the value of $\kappa$ depends on the unit of constraints. Throughout the paper, we measure model size in terms of MB (megabytes) and computation cost in terms of GBitOPs (billion of BitOPs). In this way, the desired resource constraint can be reached in the joint optimization of model parameters and bit-widths. Note that the recent concurrent work~\cite{bitprune} adopts a similar approach for mixed precision quantization with L1 regularzation on bit-widths for weights and activations, while here we explicitly define the loss as a function of computational cost in BitOPs or model size in Bytes and incorporate the target constraint into the loss directly.

\subsection{Finetuning with mixed precision}
After searching, the bit-widths of the model are still continuous values. We discretize the bit-widths with a threshold value to make the resulted network meeting the resource constraints. Specifically, we use binary search to find a threshold for both weight and activation bit-widths, which makes the resource cost of the model to be within $1\%$ deviation of the target constraint. This way, each layer or each kernel has its individual bit-widths for weights and activations learned in the previous stage, and the training enters the finetuning stage to only update model weights. The ratio between training epochs allocated to searching and finetuning is a hyper-parameter that can be freely specified. In practice, we assign $80\%$ of training epochs to searching and $20\%$ to finetuning. Here we want to emphasize that the combination of searching and finetuning constitutes the whole training procedure, and the total number of epochs of the two stages is the same as a traditional quantization-aware training procedure. 
Also, the whole procedure for updating learned parameters and scheduling hyper-parameter (learning rate, weight decay, etc.) is smooth, and do not need any re-initialization for the finetuning. Thus, our training method is one-shot, without extra retraining steps.

\subsection{Kernel-wise mixed precision quantization}
\label{sec:channel}
As mentioned above, our algorithm is not restricted to layer-wise quantization, but also supports kernel-wise quantization. Here, one kernel means weight parameters associated with a convolution filter to produce a single-channel feature map. Weight kernels in a convolution layer are assigned with different bit-width parameters $\lambda_{w_i}$, where $i$ is the index of the weight kernel. For each convolution operation of one weight kernel with the input tensor, the input tensor can also be assigned with different bit-widths. However, quantizing the input tensor with different bit-widths for different weight kernels requires large computation overhead. Here we assign the same bit-width $\lambda_a$ on the input tensor for computation with all the weight kernels. Note that~\cite{autoq} adopted the same strategy for kernel-wise quantization. For a 2D convolution layer, the number of BitOPs associated with the fractional bit-width is given by $\sum_{i}{\lambda_{w_i}\lambda_a c_{in}k_x k_y o_x o_y}$, and model size can be represented by $\sum_{i}{\lambda_{w_i} c_{in} k_x k_y}$.

%% file: figs/fig1.tex
\begin{figure*}[tb]
\centering
\includegraphics[width=0.84\linewidth]{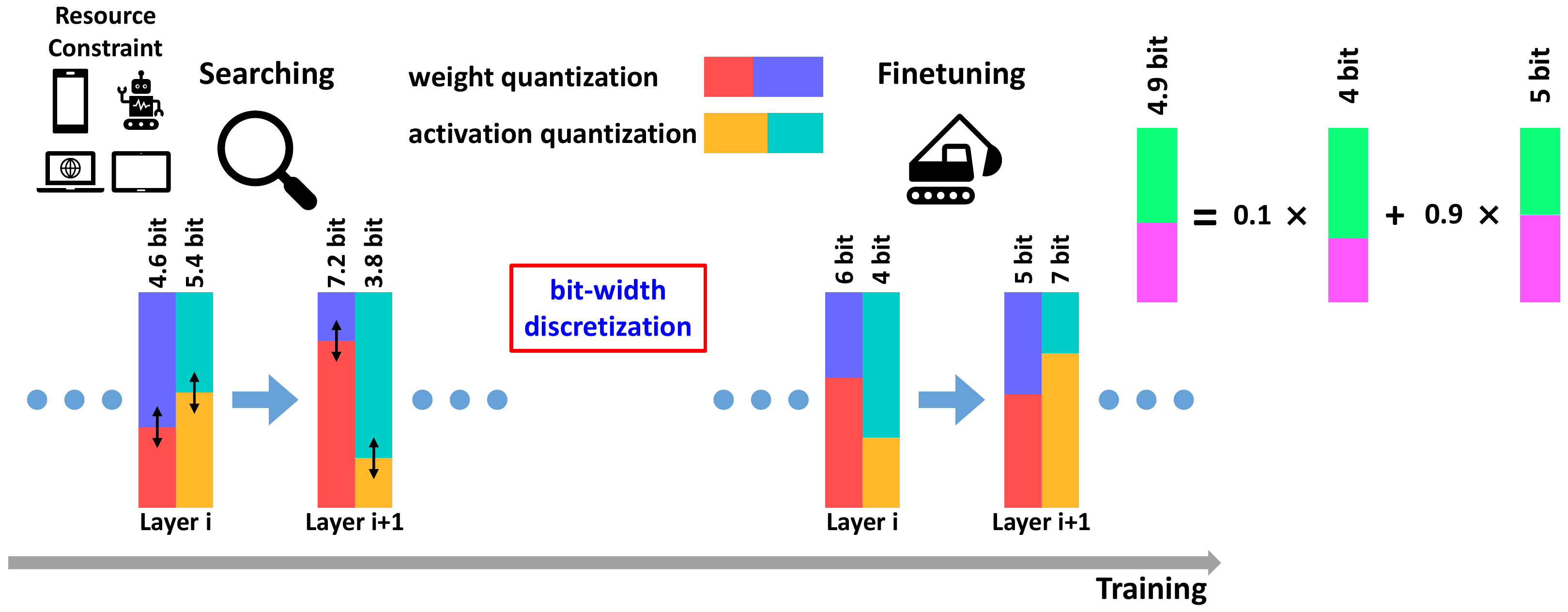}
\caption{Our differentiable bit-width searching method consists of two stages: searching with fractional bit-width and finetuning with mixed bit-width quantization.}
\label{fig:fracbit}
\end{figure*}

%% file: tex_files/experiments.tex
\section{Experiments}
In this section, we conduct quantitative experiments using FracBits and compare it with previous quantization approaches including uniform quantization algorithms PACT~\cite{pact}, LQNet~\cite{lq}, SAT~\cite{sat} and mixed precision quantization algorithms HAQ~\cite{haq}, AutoQ~\cite{autoq}, DNAS~\cite{dnas}, US~\cite{uniform_sampling}, DQ~\cite{dq}. We first compare our method with previous approaches on layer-wise mixed precision quantization. 
Then we compare our method with a previous kernel-wise mixed precision method AutoQ on kernel-wise precision search. 

\subsection{Implementation details}
We build our algorithms based on a recent quantization 
algorithm SAT~\cite{sat}, which is an improved version of PACT algorithm~\cite{pact}. PACT jointly learns quantized weights and activations where weights are quantized using the DoReFa scheme~\cite{dorefa}, while 
SAT modifies PACT with gradient calibration and scale adjusting.  
$\kappa$ is a critical parameter for the proper convergence of the network towards required resource constraints. Models under mild or aggressive constraints may couple with different values of $\kappa$. Different types of resource constraints (computational cost and model size) have different scales and requires different scales of the regularization term. However, in our experiments, we find our algorithm is not very sensitive to values of $\kappa$. We set $\kappa$ to $0.1$ for all computation cost constrained experiments, and $1$ for all model size constrained experiments. We also find it beneficial to initialize the model at some point close to the target resource constraint, facilitating more exploration close to the target model spaces. We set the initial value of $\kappa$ to $b_{t} + 0.5$ in each layer for all  experiments, where $b_t$ is the bit-width achieving similar resource constraints in the corresponding uniformly quantized model. For all channel-wise quantization experiments with both weights and activations quantized, we set the candidate bit-widths to be 2-8. For all other experiments including weight-only quantization and kernel-wise quantization, we set the candidate bit-widths to be 1-8. Since the first and the last layers in a neural network have crucial impact on the performance of the model, we fix the bit-width of the first and last layer to 8 bit following~\cite{sat}.

For all experiments, we use cosine learing rate scheduler without restart. Learning rate is initially set to
0.05 and updated every iteration for totally 150 epochs. We use SGD optimizer
with a momentum weight of 0.9 without damping, and weight decay of $4\times10^{-5}$. The batch size
is set to 2048 for all models. The warmup strategy
suggested in~\cite{warmup} is also adopted by linearly increasing the learning rate every iteration to $batch size/256 \times 0.05$ for the first five epochs before using the cosine annealing scheduler. Bit-width search is conducted in the first 120 epochs after the warmup stage. At the 121th epoch, all fractional bit-width will be discretized to integer bits, and the network will be further finetuned for the rest 30 epoches. We do not observe any glitch in the training loss in this discretization process, potentially due to the insignificant difference in quantized values of two neighboring bit-widths. 

\input{tabs/tab2}

\input{tabs/tab3}

\input{figs/fig2}

\subsection{Quantization with layer-wise precision}\label{sec:exp_layerwise}
We compare FracBits with previous quantization algorithms on layer-wise precision search. We conducted experiments on MobileNet V1/V2 and ResNet18. Since FracBits can be used for both computation cost constrained and model size constrained bit-width search, we conduct experiments on both settings to validate the effectiveness of our approach.

Table~\ref{table:comp_mobilenet} shows experiment results of layer-wise computation cost constrained quantization on MobileNet V1/V2. 
Previous methods HAQ~\cite{haq} and AutoQ~\cite{autoq} use PACT as quantization scheme, while DQ uses a similar scheme to PACT with learnable clipping bounds. 
Derived from PACT, SAT is a strong uniform quantization baseline which already outperforms all previous mixed precision methods. For example, it already achieves 71.3\% on 4-bit MobileNet V1 and 70.8\% on 4-bit MobileNet V2, almost closing the gap between full precision models (71.7\% for MobileNet V1 and 71.8\% for MobileNet V2) and quantized ones. We believe that validating the effectiveness of our FracBits algorithm based on SAT is helpful towards seeking the limit of mixed precision quantization algorithms. We find that FracBits-SAT achieves slightly better performance compared to SAT on 4-bit MobileNet V1/V2, and achieves significantly better result on 3-bit models, which proves its effectiveness on strong uniform quantization baselines. It has a $1.6\%$ absolute gain on 3-bit MobileNet V1 and a $0.6\%$ gain on MobileNet V2\footnote{In MobileNet V2, some convolution layers are not followed by ReLU activation. Here we use 
double-sided quantization for outputs of these layers, meaning that they are clipped into an interval of $[-\alpha, \alpha]$. Our re-implemented SAT also adopts such double-side clipping.}
under the same computation cost budget. 

We show comparison with more algorithms on ResNet18, as listed in Table~\ref{table:resnet}. Here we compare with uniform precision approaches PACT, LQNet and mixed precision approaches DNAS, DQ, AutoQ, and US. Except DQ, all mixed precision approaches use PACT as quantization scheme. Since all methods report different accuracies for full precision (FP) models, we also add the top-1 accuracy of FP models reported in corresponding papers and report the relative accuracy drop for each method. Here we also include results of directly applying our method on PACT scheme, and denote this as FracBits-PACT. Comparing absolute accuracy, FracBits-PACT achieves comparable performance as state-of-the-art mixed precision methods. 
Note DNAS uses several tricks in training to boost performance, thus its result is not directly comparable to others. Comparing relative accuracy drop, our method achieves least performance drop on 3-bit ResNet18. Enhanced by SAT quantization method, FracBits-SAT further improves over SAT baseline and achieves only $0.8\%$ accuracy drop on 3-bit ResNet18 and even a $0.4\%$ performance gain on 4-bit ResNet18. Note DQ and our method are one-shot differentiable method which only need one pass of training to obtain the final model, and are much more efficient than the other mixed quantization approaches (DNAS, AutoQ, US).

To have a more intuitive understanding of the learned bit-width structure from our algorithm, we plot the bit-widths from different layers for 3-bit MobileNet V2 and ResNet18, as shown in Fig.~\ref{fig:layerwise}. We find that models for mixed quantization contrained on computational cost generally uses more bit-width on the late stage of the network, potentially due to the larger computation cost of  early layers than later layers. Also, in MobileNet V2, depth-wise convolutions result in more bit-width than point-wise convolutions due to their low computation cost.

\input{tabs/tab4}

\input{figs/fig3}

For model size constrained quantization, we show comparison with previous methods Deep Compression~\cite{han2015deep}, HAQ and uniform quantization approach SAT in Table~\ref{table:size_mobilenet}. Our FracBits-SAT outperforms mixed precision methods HAQ and strong uniform quantization baseline SAT on all experimented bit-widths consistently. Note that FracBits has an over $3\%$ absolute gain on top-1 accuracy over SAT on 2-bit MobileNet V1/V2. On the challenging 3-bit setting where quantized models already achieve similar performance as full precision ones, FracBits also outperforms SAT with a $0.6\%$ margin on MobileNet V1 and a $0.8\%$ gain on MobileNet V2 in top-1 accuracy.

\input{tabs/tab5}

\subsection{Quantization with kernel-wise precision}\label{sec:exp_channelwise}
In this section, we experiment with quantization with kernel-wise precision. Among previous approaches, only AutoQ~\cite{autoq} has experiments on kernel-wise precision which we will compare with. In Table~\ref{table:chwise_compare}, we denote kernel-wise FracBits based on PACT and SAT as FB-PACT-K and FB-SAT-K, and compare them with AutoQ and uniform precision method SAT. FB-PACT-K achieves 
comparable results as AutoQ on MobileNet V2 and ResNet18, while being much more efficient than the RL based method which needs to train hundreds of model variants, thanks to the differentiable formulation. FB-SAT-K outperforms SAT significantly with $1.0\%$  and $0.8\%$ 
increase on top-1 accuracy on 3 and 4
-bit MobileNet V2, respectively,
and with $0.5\%$
increase on top-1 accuracy on both 3 and 4
-bit ResNet18. 
Compared to layer-wise precision models, FB-SAT-K outperforms FracBits-SAT by $0.4\%$ and $0.3\%$ 
on 3 and 4-bit MobileNet V2, respectively. 
It also outperforms layer-wise FracBits-SAT by $0.4\%$ on 3-bit ResNet18, proving our kernel-wise quantization method can further improve over strong layer-wise mixed-precision models. Fig.~\ref{fig:kernelwise} illustrates the bit-width distribution against layer indices for 3-bit MobileNet V2 and ResNet18. We can see that 3-bit MobileNet V2 assign low bits in the early layers and intermediate bottleneck layers, while 3-bit ResNet-18 assign low bits only in early layers. We believe that the point-wise convolutions in MobileNet V2 have much larger computation cost compared to depth-wise convolutions and thus they receive a larger resource penalty during optimization, leading to more compression by lower bit-widths.



%% file: tabs/tab2.tex
\begin{table}[tb]
\centering
\footnotesize
\setlength{\tabcolsep}{1.6pt}
\caption{Comparison of computation cost constrained layer-wise quantization of our method and previous approaches on ImageNet with MobileNet V1/V2. Note that accuracies are in \% and bitops are in B (billion).}
\begin{tabular}{|l|l|c|c|c|c|c|c|}
\hline
\multirow{2}{*}{}             & bit-width     & \multicolumn{3}{c|}{3} & \multicolumn{3}{c|}{4} \\ \cline{2-8} 
                              & method        & top-1 & top-5 & bitops & top-1 & top-5 & bitops  \\ \hline
\multirow{3}{*}{MBNetV1} &  HAQ          & -     & -     & -      & 67.4  & 87.9  & -        \\ \cline{2-8} 
                              &  SAT           & 67.1  & 87.1  & 5.73   & 71.3  & 89.9  & 9.64    \\ \cline{2-8} 
                              & FracBits-SAT  & \textbf{68.7}  & \textbf{88.2}  & 5.78   & \textbf{71.4}  & \textbf{90.0}  & 9.63    \\ \hline
\multirow{5}{*}{MBNetV2} & HAQ          & -     & -     & -      & 67    & 87.3  & -          \\ \cline{2-8}  
                              & AutoQ         & -     & -     & -      & 69    & 89.4  & -    \\ \cline{2-8} 
                              & DQ            & -     & -     & -      & 69.7  & -     & -         \\ \cline{2-8} 
                              &  SAT        & 67.2  & 87.3    & 3.32   & 70.8  & 89.7  & 5.35   \\ \cline{2-8} 
                              & FracBits-SAT  & \textbf{67.8}  & \textbf{87.6}  & 3.33   & \textbf{71.3}  & \textbf{90.1}  & 5.35   \\ \hline
\end{tabular}
\label{table:comp_mobilenet}
\end{table}

%% file: tabs/tab3.tex
\begin{table*}[tb]
\centering
\footnotesize
\caption{Comparison of computation cost constrained layer-wise quantization of our method and previous approaches on ImageNet with ResNet18. Note bitops of US~\cite{uniform_sampling} and DNAS~\cite{dnas} does not include first and last layer in their papers, and US shows different bitops numbers from ours. We give an estimation of their bitops based on the difference with uniformly quantizated models. Note that accuracies are in \% and bitops are in B (billion).}
\begin{tabular}{|l|c|c|c|c|c|c|c|}
\hline
bit-width     & \multicolumn{3}{c|}{3}        & \multicolumn{3}{c|}{4}        & FP    \\ \hline
method        & top-1 & $\Delta_{acc}$ & bitops & top-1 & $\Delta_{acc}$ & bitops & top-1 \\ \hline
PACT~\cite{pact}          & 68.3  & -1.9         & 22.83  & 69.2  & -1.0           & 34.70   & 70.2  \\ \hline
LQNet~\cite{lq}         & 68.2  & -2.1         & 22.83  & 69.3  & -1.0           & 34.70   & 70.3  \\ \hline
DNAS~\cite{dnas}          & 68.7  & -2.3         & 24.34*  & 70.6  & -0.4         & 35.17*  & 71.0    \\ \hline
DQ~\cite{dq}            & -     & -            & -      & 70.1  & \textbf{-0.2}         & -      & 70.3  \\ \hline
AutoQ~\cite{autoq}         & -     & -            & -      & 68.2  & -1.7         & -      & 69.9  \\ \hline
US~\cite{uniform_sampling}            & 69.4  & -1.5         & 22.11* & 70.5  & -0.4         & 33.74* & 70.9  \\ \hline
FracBits-PACT & 69.1    & \textbf{-1.1}         & 22.70  & 69.7  & -0.5         & 34.73  & 70.2  \\ 
\hhline{|=|=|=|=|=|=|=|=|}
SAT~\cite{sat}	 & 69.3	& -0.9	& 22.83	& 70.3	&0.1	&34.70	&70.2 \\ \hline
FracBits-SAT &	69.4 &	\textbf{-0.8} & 22.93	&70.6	& \textbf{0.4}	&34.70	&70.2 \\ \hline
\end{tabular}
\label{table:resnet}
\end{table*}

%% file: figs/fig2.tex
\begin{figure}[!htb]
\centering
\begin{subfigure}[!htb]{.41\textwidth}
\centering
   \includegraphics[width=\linewidth]{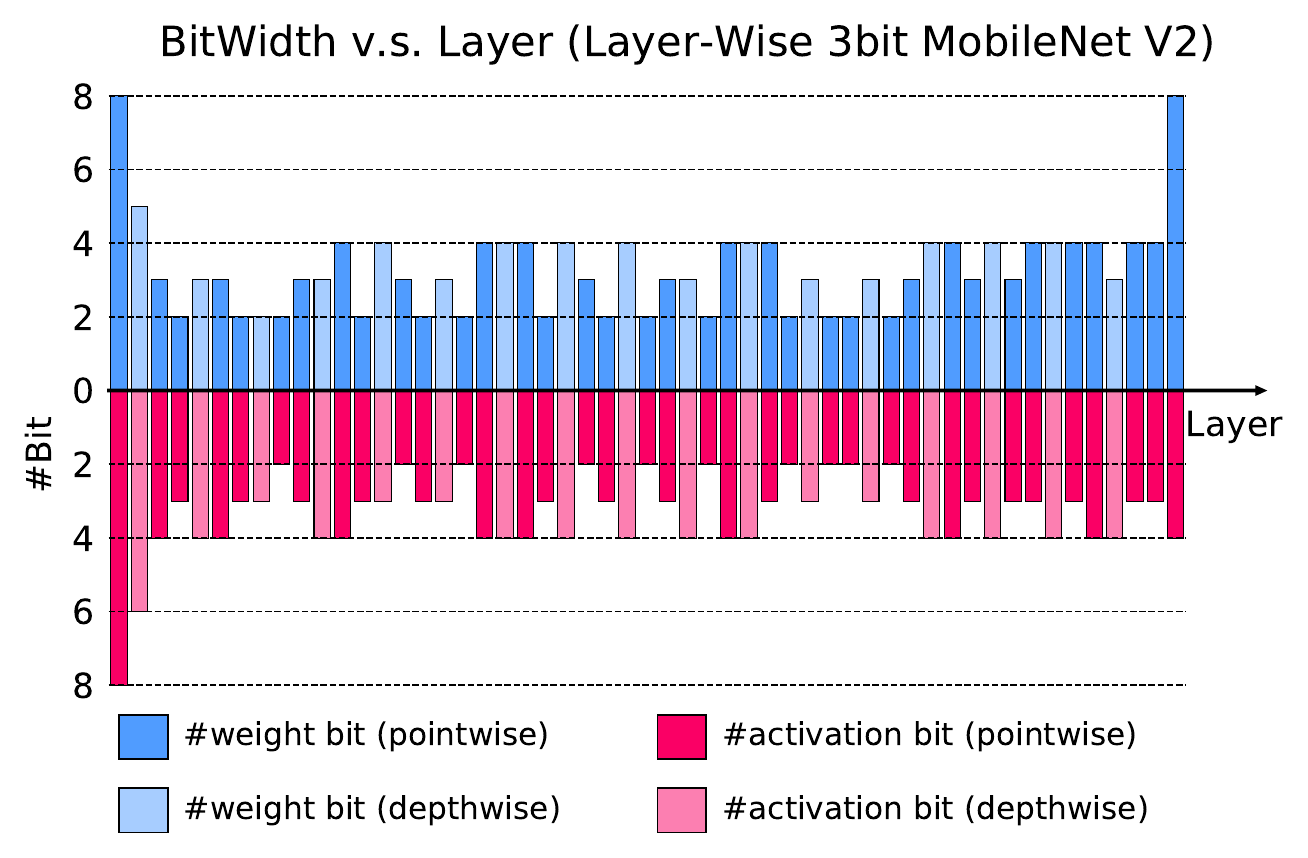}
   \caption{}
   \label{fig:mbv2_layerwise}
\end{subfigure}
\begin{subfigure}[!htb]{.41\textwidth}
\centering
   \includegraphics[width=\linewidth]{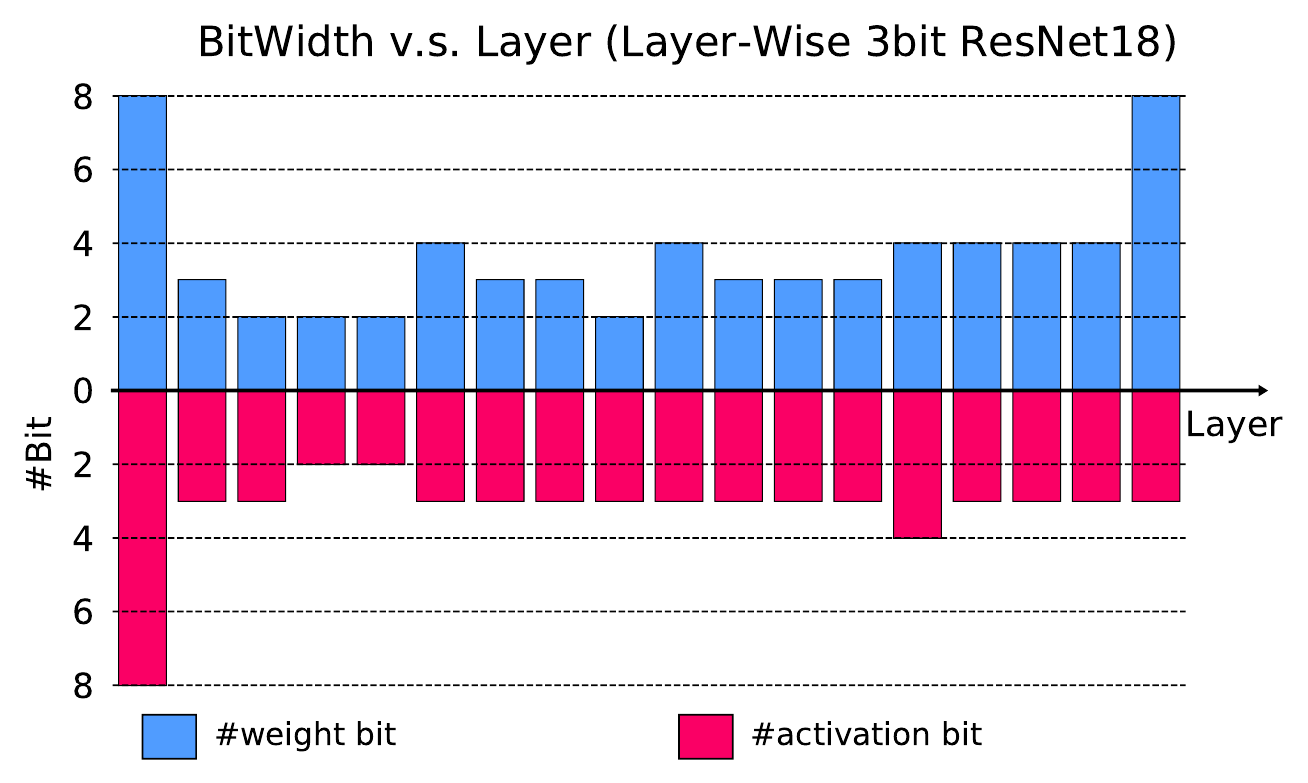}
   \caption{}
   \label{fig:res18_layerwise}
\end{subfigure}
   \caption{Layer-wise mixed precision quantization for 3 bit MobileNet V2 (a) and ResNet18 (b).}
\label{fig:layerwise}
\end{figure}

%% file: tabs/tab4.tex
\begin{table}[tb]
\centering
\footnotesize
\setlength{\tabcolsep}{2pt}
\caption{Comparison of model size constrained layer-wise quantization of our method and previous approaches on ImageNet with MobileNet V1/V2. Note that accuracies are in \% and sizes are in MB.}
\begin{tabular}{|l|l|c|c|c|c|c|c|}
\hline
\multirow{2}{*}{}             & bit-width    & \multicolumn{3}{c|}{2}               & \multicolumn{3}{c|}{3}                            \\ \cline{2-8} 
                              & method       & top-1         & top-5         & size & top-1         & top-5         & size\\ \hline
\multirow{4}{*}{MBNetV1} & DeepComp     & 37.6          & 64.3          & -    & 65.9          & 86.9          & -       \\ \cline{2-8} 
                              & HAQ         & 57.1          & 81.9          & -    & 67.7          & 88.2          & -      \\ \cline{2-8} 
                              & SAT          & 66.3          & 86.8          & 1.83 & 70.7          & 89.5          & 2.22  \\ \cline{2-8} 
                              & FracBits-SAT & \textbf{69.7} & \textbf{88.9} & 1.81 & \textbf{71.3} & \textbf{90.0} & 2.23  \\ \hline
\multirow{4}{*}{MBNetV2} & DeepComp    & 58.1          & 82.2          & -    & 68.0            & 88.0            & -      \\ \cline{2-8} 
                              & HAQ          & 66.8          & 87.3          & -    & 70.9          & 89.8          & -      \\ \cline{2-8} 
                              & SAT         & 66.8          & 87.2          & 1.83 & 71.1          & 89.9          & 2.11  \\ \cline{2-8} 
                              & FracBits-SAT & \textbf{69.9} & \textbf{89.3} & 1.84 & \textbf{71.9} & \textbf{90.4} & 2.12  \\ \hline
\end{tabular}
\label{table:size_mobilenet}
\end{table}

%% file: figs/fig3.tex
\begin{figure}[!htb]
\centering
\begin{subfigure}[!htb]{.44\textwidth}
\centering
   \includegraphics[width=\linewidth]{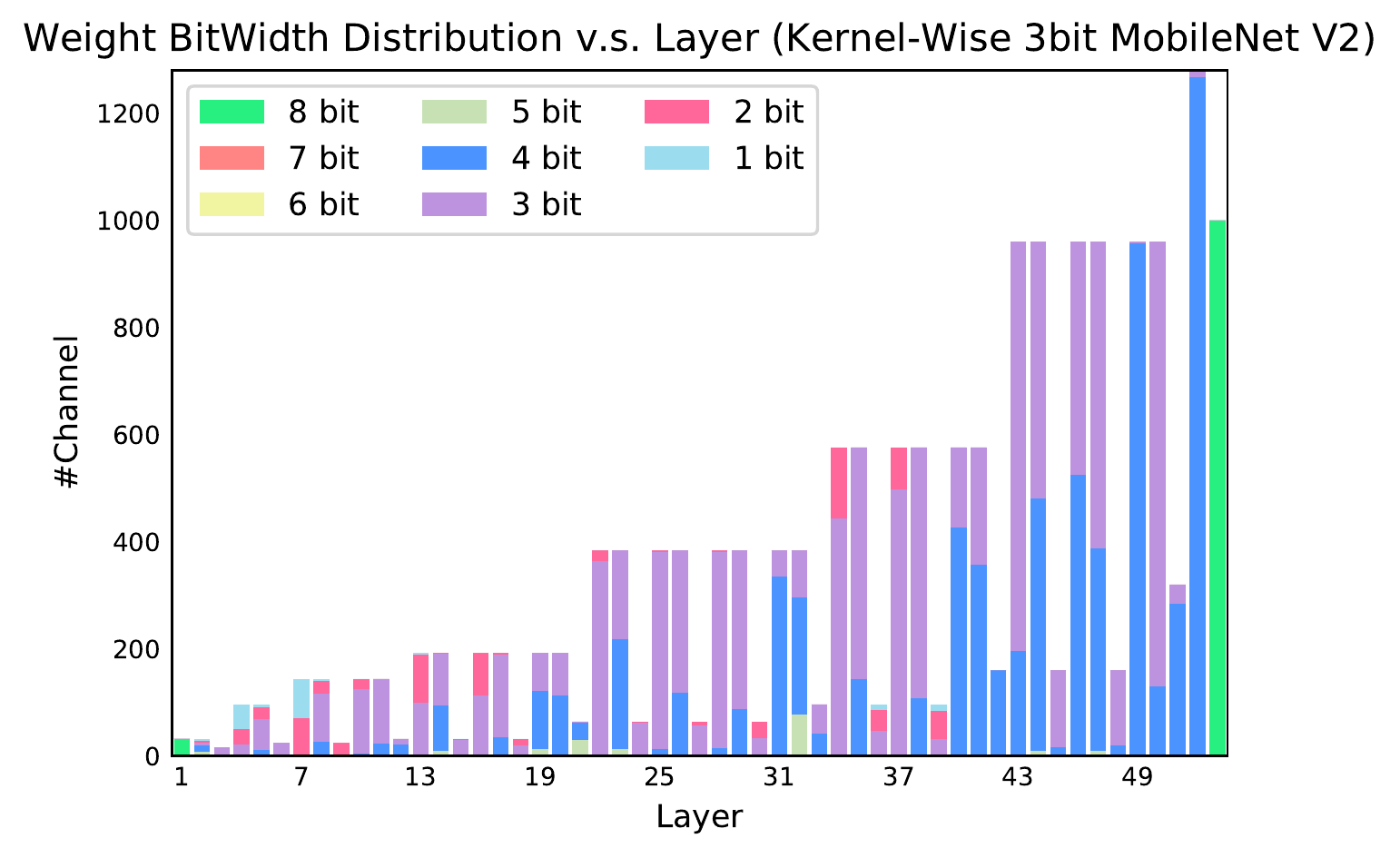}
   \caption{}
   \label{fig:mbv2_kernelwise}
\end{subfigure}
\begin{subfigure}[!htb]{.44\textwidth}
\centering
   \includegraphics[width=\linewidth]{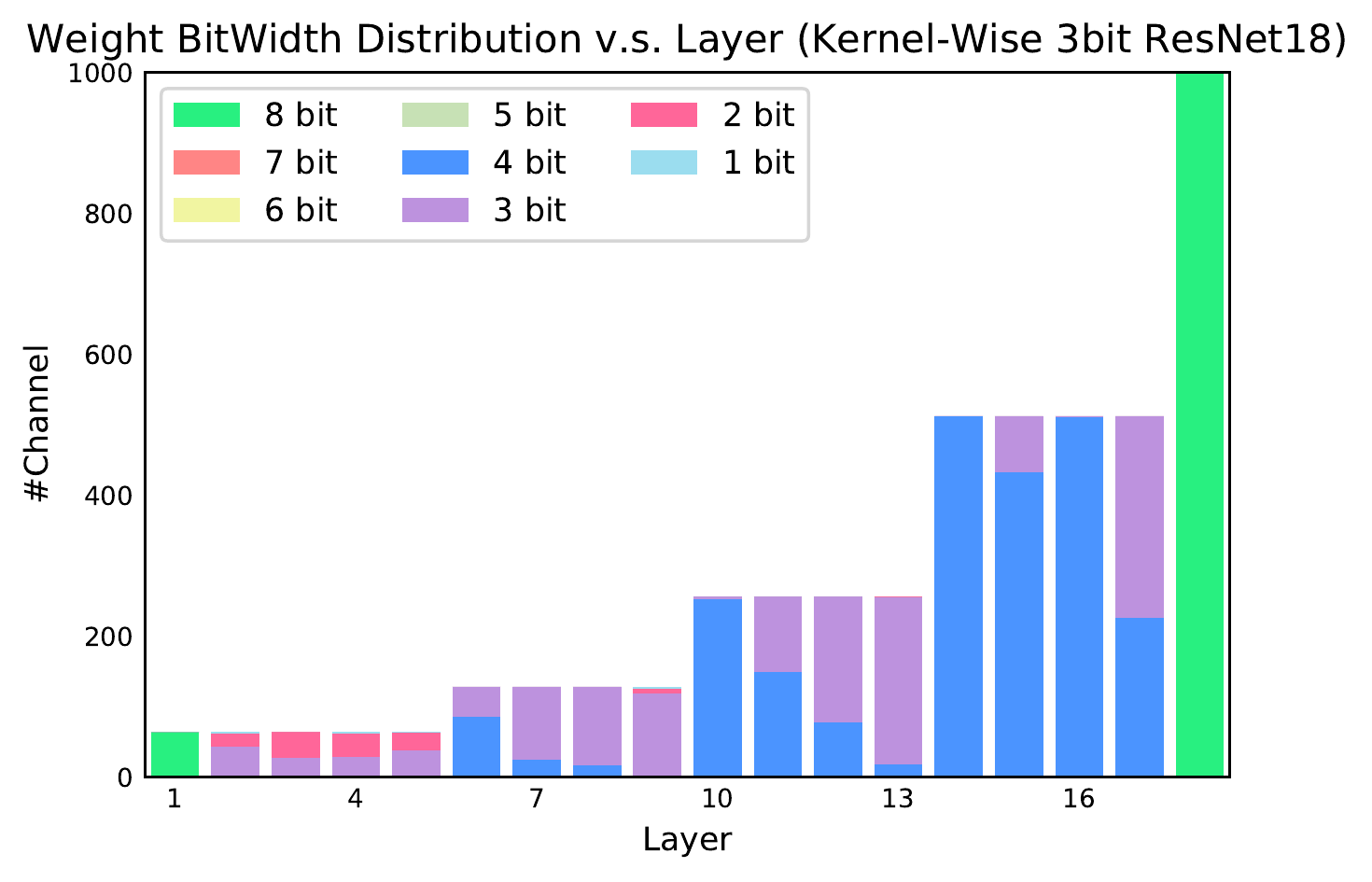}
   \caption{}
   \label{fig:res18_kernelwise}
\end{subfigure}
   \caption{Kernel-wise mixed precision quantization for 3 bit MobileNet V2 (a) and ResNet18 (b).}
\label{fig:kernelwise}
\end{figure}

%% file: tabs/tab5.tex
\begin{table}[tb]
\centering
\footnotesize
\setlength{\tabcolsep}{2pt}
\caption{Comparison of computation cost constrained kernel-wise quantization of our method and previous approaches on MobileNet V2 and ResNet18. Note that accuracies are in \% and bitops are in B (billion). FB-PACT-K and FB-SAT-K denote FracBits for kernel-wise quantization with PACT and SAT quantization schemes, respectively. }
\scalebox{1.0}{
\begin{tabular}{|c|l|c|c|c|c|c|c|}
\hline
\multirow{2}{*}{}             & bit-width       & \multicolumn{3}{c|}{3}                 & \multicolumn{3}{c|}{4}                 \\ \cline{2-8} 
                              & method          & top-1         & top-5         & bitops & top-1         & top-5         & bitops \\ \hline
\multirow{4}{*}{MBNetV2} & AutoQ          & -             & -             & -      & 70.8          & \textbf{90.3} & -      \\ \cline{2-8} 
                              & FB-PACT-K & 68.0 & 87.8 & 3.33   & 70.9 & 89.5          & 5.36   \\ \cline{2-8}
                              & SAT             & 67.2          & 87.3          & 3.32   & 70.8          & 89.7          & 5.35   \\ \cline{2-8} 
                              & FB-SAT-K  & \textbf{68.2} & \textbf{87.9} & 3.35   & \textbf{71.6} & 90.0 & 5.33   \\ \hline
\multirow{4}{*}{ResNet18}     & AutoQ           & -             & -             & -      & 69.8          & 88.4          & -      \\ \cline{2-8} 
                              & FB-PACT-K & 69.0 & 88.3 & 23.01  & 69.9 & 88.8 & 34.70  \\ \cline{2-8} 
                              & SAT            & 69.3          & \textbf{88.9} & 22.83  & 70.3          & 89.5          & 34.70  \\ \cline{2-8} 
                              & FB-SAT-K  & \textbf{69.8} & \textbf{88.9 }         & 22.87  & \textbf{70.8} & \textbf{89.6} & 34.82  \\ \hline
\end{tabular}
}
\label{table:chwise_compare}
\end{table}

%% file: tex_files/conclusion.tex
\section{Conclusion}
We propose a new formulation named FracBits for mixed precision quantization. We formulate the bit-width of each layer or kernel with a continuous learnable parameter that can be instantiated by interpolating quantized parameters of two neighboring bit-widths. Our method facilitates differentiable optimization of layer-wise or kernel-wise bit-width in a single shot of training.
With only a regularized term to penalize extra computational resource in the training process, our method is able to discover proper bit-width configurations for different models, outperforming previous mixed precision and uniform precision approaches. We believe our method will motivate research along low-precision neural networks, and low-cost computational models.

%% file: tex_files/supp.tex
\section{Ablation Study}

We show some ablation study related to our method in this section. Since our framework is clean and only involves one hyper-parameter $\kappa$, we show a comparative study of using different values of $\kappa$. Another variant we can compare with is using stochastic bit-width following~\cite{dnas} instead of determined fractional bits in the searching stage. Towards this end, we utilize gumbel softmax~\cite{jang2016gumbel} to generate stochastic bit-widths based on the original fractional bit-widths. $\kappa$ is set to the same value as the deterministic approach and temperature for gumbel softmax is set to 1. The results are show in Table~\ref{ablation}. With smaller value of $\kappa$ at 0.05, FracBits-SAT shows worse performance than the original model on 3-bit MobileNet V1, indicating small values of $\kappa$ may fail to enforce the model to explore within the computational constraint due to weak penalty. With larger value of 0.2, the models still perform similarly as with $\kappa=0.1$, proving the robustness of our method within a proper range of $\kappa$. We have also experimented with $\kappa$ as large as $0.5$ which results in a rapid descend of bit-widths values in the beginning of training and generates poor result. With gumbel softmax, the result is significantly worse than the original FracBits-SAT, proving the advantage of our approach.

\input{tabs/tab6}

%% file: tabs/tab6.tex
\begin{table}[!htb]
\centering
\footnotesize
\setlength{\tabcolsep}{3pt}
\caption{A comparative study of our method with different configurations and hyper-parameters on MobileNet V1 for compution cost constrained quantization. w/ gumbel denotes the models using gumbel softmax to sample stochastic bit-width in searching. Note that accuracies are in \% and bitops are in B (billion).}
\begin{tabular}{|l|c|c|c|c|c|c|}
\hline
bit-width    & \multicolumn{3}{c|}{3} & \multicolumn{3}{c|}{4}  \\ \hline
             & top-1 &top-5 & bitops & top-1 & top-5 & bitops  \\ \hline
FracBits-SAT & 68.7  & 88.2 & 5.78    & 71.4  &  90.0 & 9.63    \\ \hline
w/ gumbel    &  65.5        & 86.1 & 5.73      & 69.8 & 89.0    & 9.62  \\ \hline
$\kappa$=0.2    & 68.9   & 88.1 & 5.78   & 71.4   & 89.9 &9.67   \\ \hline
$\kappa$=0.05   & 67.4   & 87.4 &5.73    & 70.5  &89.4   &9.68    \\ \hline
\end{tabular}
\label{ablation}
\end{table}

